\documentclass[a4paper,10pt]{article}
\usepackage[dvips]{graphicx}
\usepackage{theorem}
\usepackage{subeqn}
\usepackage{floatflt}
\newcommand{\gbf}[1] {\mbox{\boldmath${#1}$\unboldmath}}
\newcommand{\be}{\begin{equation}}
\newcommand{\ee}{\end{equation}}
\newcommand{\beq}{\begin{equation}}
\newcommand{\eeq}{\end{equation}}
\newcommand{\bed}{\begin{displaymath}}
\newcommand{\eed}{\end{displaymath}}
\newcommand{\beqa}{\begin{eqnarray}}
\newcommand{\eeqa}{\end{eqnarray}}
\newcommand{\beqann}{\begin{eqnarray*}}
\newcommand{\eeqann}{\end{eqnarray*}}
\newcommand{\bseq}{\begin{subequations}}
\newcommand{\eseq}{\end{subequations}}

\newcommand{\ba}{\begin{array}}
\newcommand{\ea}{\end{array}}

\newcommand{\M}{{\bf M}}

\newcommand{\negr}[1]{{\bf {#1}}}

 \makeatletter
\theoremstyle{plain}

\def\@normalsize{\@setsize\normalsize{10pt}\xpt\@xpt
\abovedisplayskip 3pt plus 1pt minus 3pt
\belowdisplayskip\abovedisplayskip \abovedisplayshortskip \z@ plus
3pt \belowdisplayshortskip 3pt plus 2pt minus 3pt
\let\@listi\@listI}
\makeatother

\begin{document}
\date{}  
\title {\noindent\bf A Comparative Study
of Parallel Kinematic Architectures for Machining Applications}
\author{\begin{tabular}[t]{c}
 {Philippe Wenger$^1$, Cl\'ement Gosselin$^2$ and Damien Chablat$^1$} \\
 {\em $^1$Institut de Recherche en Communications et Cybern\'etique de Nantes
 \footnote{IRCCyN: UMR n$^\circ$ 6597 CNRS, \'Ecole Centrale de Nantes,
                     Universit\'e de Nantes, \'Ecole des Mines de Nantes}} \\
 {\em 1, rue de la No\"e, 44321 Nantes, France} \\
 {\em $^2$D\'epartement de g\'enie m\'ecanique, Universit\'e Laval} \\
 {\em Qu\'ebec, Qu\'ebec, Canada, G1K 7P4} \\
 {\bf Philippe.Wenger@irccyn.ec-nantes.fr}
\end{tabular}}
\maketitle
{\noindent\bf Abstract:} {Parallel kinematic mechanisms are interesting  alternative designs for machining applications. Three 2-DOF parallel mechanism architectures dedicated to machining applications are studied in this paper. The three mechanisms have two constant length struts gliding along fixed linear actuated joints with different relative orientation. The comparative study is conducted on the basis of a same prescribed Cartesian workspace for the three mechanisms. The common desired workspace properties are a rectangular shape and given kinetostatic performances. The machine size of each resulting design is used as a comparative criterion. The 2-DOF machine mechanisms analyzed in this paper can be extended to 3-axis machines by adding a third joint.
\section{Introduction}
Most industrial 3-axis machine tools have a PPP kinematic
architecture with orthogonal joint axes along the x, y, z
directions. Thus, the motion of the tool in any of these direction
is linearly related to the motion of one of the three actuated
axes. Also, the performances (e.g. maximum speeds, forces,
accuracy and rigidity) are constant in the most part of the
Cartesian workspace, which is a parallelepiped. In contrast, the
common features of most existing PKM (Parallel Kinematic Machine)
are a Cartesian workspace shape of complex geometry and highly non
linear input/output relations. For most PKM, the Jacobian matrix
which relates the joint rates to the output velocities is not
constant and not isotropic. Consequently, the performances may
vary considerably for different points in the Cartesian workspace
and for different directions at one given point, which is a
serious drawback for machining applications \cite{Treib:1998}. To
be of interest for machining applications, a parallel kinematic
architecture should preserve good workspace properties (regular
shape and acceptable kinetostatic performances throughout). It is
clear that some parallel architectures are more appropriate than
others, as it has already been shown in previous studies
\cite{Wenger:1999,Kim:1997}. The aim of this paper is to compare
three parallel kinematic architectures. To limit the analysis, the
study is conducted for 2-DOF mechanisms but the results can be
extrapolated to 3-DOF architectures. The three mechanisms studied
have two constant length struts gliding along fixed linear
actuated joints with different relative orientation. Each
mechanism is defined by a set of three design variables. Given a
prescribed Cartesian rectangular region with given kinetostatic
performances, we calculate the link dimensions and joint ranges of
each mechanism for which the prescribed region is included in a
t-connected region of the mechanism and the kinetostatic
constraints are satisfied. Then, we compare the size of the
resulting mechanisms. The organisation of this paper is as
follows. The next section presents the mechanism studied. Section
3 is devoted to the comparison of three architectures. The last
section concludes this paper.
\section{Kinematic study}
\subsection{Serial Topology with Three Degrees of Freedom}
Most industrial machine tools use a simple PPP serial topology
with three orthogonal prismatic joint axes
(Figure~\ref{figure:machine_PPP}).

 \begin{figure}[hb]
    \begin{center}
        \includegraphics[width=49mm,height=36mm]{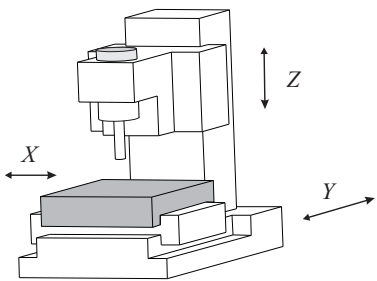}
        \caption{Typical industrial 3-axis machine-tool}
        \protect\label{figure:machine_PPP}
    \end{center}
 \end{figure}

For a $PPP$ topology, the kinematic equations are :
 \bed
  \negr J \dot{\gbf \rho} = {\bf \dot p}
  ~~~~~{\rm with} ~~~~~
  \negr J = \negr 1_{3 \times 3} \nonumber
 \eed
where ${\bf \dot p} = [x~y~z]^T$ is the velocity-vector of the
tool center point $P$ and ${\dot{\gbf \rho}} =
[\rho_1~\rho_2~\rho_3]^T$ is the velocity-vector of the prismatic
joints. The Jacobian kinematic matrix \negr J being the identity
matrix, the ellipsoid of manipulability of velocity and of force
\cite{Yoshikawa:85} is a unit sphere for all the configurations in
the  Cartesian workspace. The problem of the $PPP$ topology is
that the actuator controlling the $Y$ axis supports at the same
time the workpiece and the actuator controlling the displacement
of  the $X$ axis, which affects the dynamic performances. To solve
this problem, it is possible to use more suitable kinematic
architectures like parallel or hybrid topologies.
\subsection{The Parallel Mechanisms Studied}
We focus our study on the use of a 2-DOF parallel mechanism
(Figure~\ref{figure:PRR_RP_topology}) for the  motion of the table
of the machine tool depicted in (Figure~\ref{figure:machine_PPP}).

\begin{figure}[hbt]
    \begin{center}
    \begin{tabular}{cc}
       \begin{minipage}[t]{60 mm}
           \centerline{\hbox{\includegraphics
           [width= 62mm,height= 32mm]{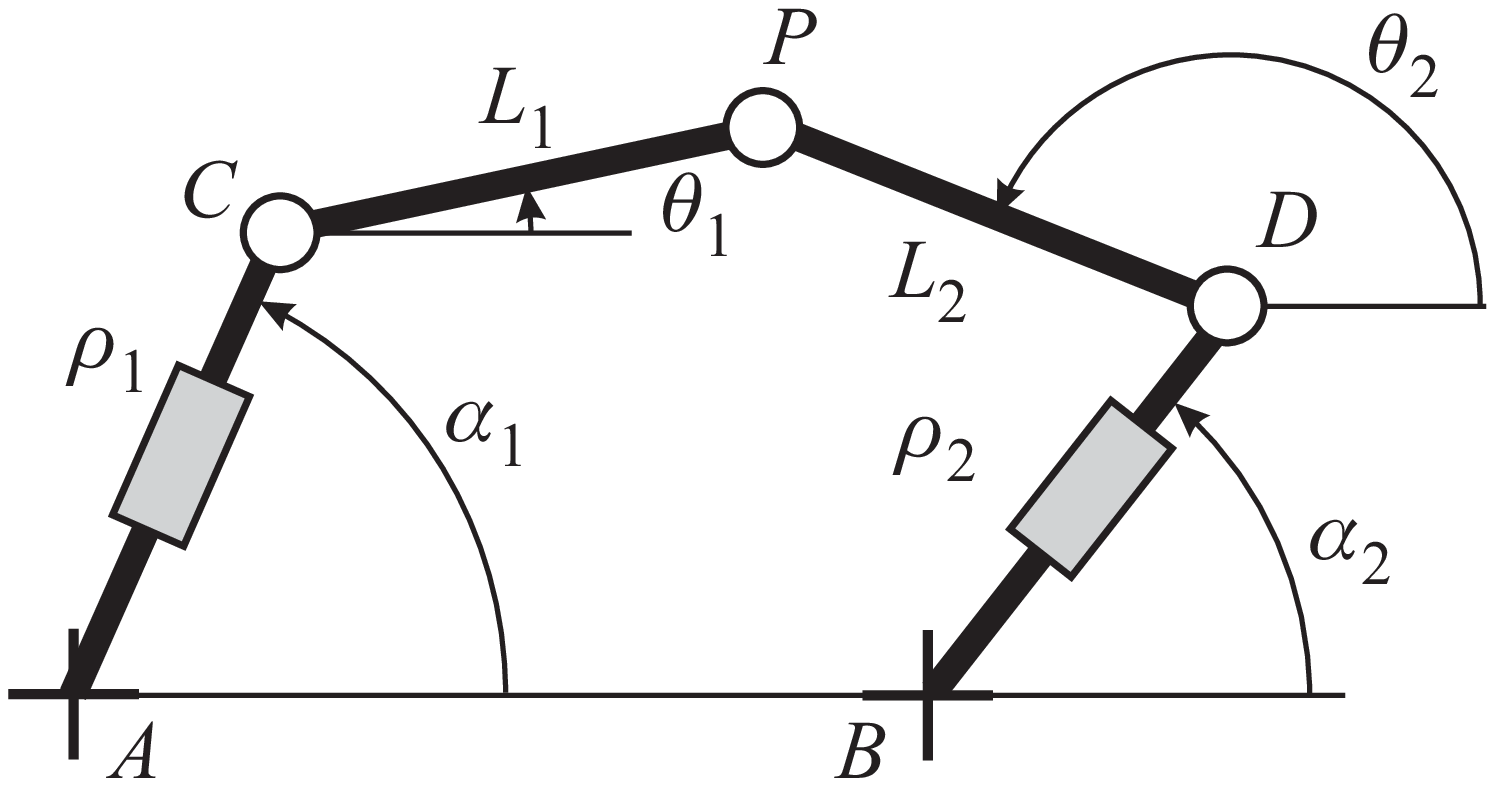}}}
           \caption{Two degree-of-freedom parallel mechanism}
           \protect\label{figure:PRR_RP_topology}
       \end{minipage} &
       \begin{minipage}[t]{60 mm}
           \centerline{\hbox{\includegraphics
           [width= 56mm,height= 37mm]{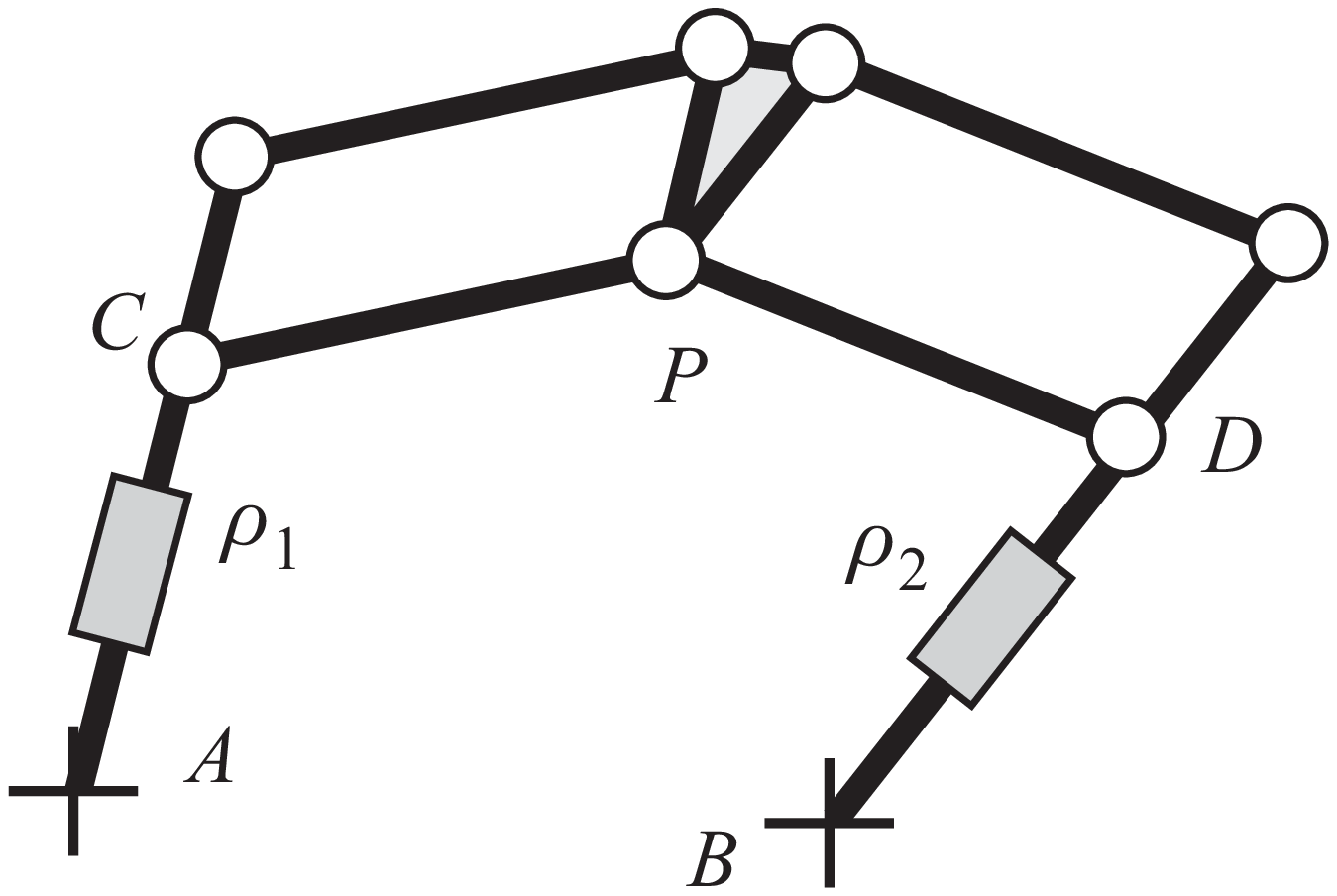}}}
           \caption{Two degree-of-freedom parallel mechanism with control
of the  orientation }
           \protect\label{figure:PRR_RP_Plus_topology}
       \end{minipage}
    \end{tabular}
    \end{center}
\end{figure}
\par
The joint variables are the variables $\rho_1$ and $\rho_2$
associated with the two actuated prismatic joints and the output
variables are the position of the tool center point $P = [x~y]^T$.
The mechanisms can be parameterized by the lengths $L_0$, $L_1$
and $L_2$, the angles $\alpha_1$ and $\alpha_2$ and the actuated
joint ranges $\Delta \rho_1$ and $\Delta \rho_2$
(Figure~\ref{figure:PRR_RP_topology}). To reduce the number of
design parameters, we impose $L_1 = L_2$ and $\Delta \rho_1 =
\Delta \rho_2$. This simplification also provides symmetry and, in
turn, reduces the manufacturing costs.
\par
To control the orientation of the reference frame attached to the
tool center point $P$, two parallelograms can be used which also
increase the rigidity of the structure
(Figure~\ref{figure:PRR_RP_Plus_topology}).
\subsection{Kinematics of the Parallel Mechanism Studied}
The velocity ${\bf \dot p}$ of $P$ can be written in two
different ways. By traversing the closed-loop $(ACP-BDP)$ in the
two possible directions, we obtain
 \bseq
 \be
  {\bf \dot p} =
  {\bf \dot c} + {\dot \theta}_1 \negr E (\negr p - \negr c)
  \label{e:MGD_1}
 \ee
and
 \be
  {\bf \dot p} =
  {\bf \dot d} + {\dot \theta}_2 \negr E (\negr p - \negr d)
  \label{e:MGD_2}
 \ee
 \eseq
where E is the rotation matrix,
 \beqa
  \negr E = \left[
              \begin{array}{cc}
                 0 & -1 \\ 1 & 0
              \end{array}
            \right]\nonumber
 \eeqa
\negr c and \negr d represent the position vector of the points
$C$ and $D$, respectively.
\par
Moreover, the velocity ${\bf \dot c}$ and  ${\bf \dot d}$ of the
points $C$ and $D$ are given by,
 \beqa
  {\bf \dot c} = \frac{\negr c - \negr a}{||\negr c - \negr a||}
                 {\dot \rho}_1 =
  \left[
    \begin{array}{c}
      \cos(\alpha_1) \\
      \sin(\alpha_1)
    \end{array}
  \right] {\dot \rho}_1
  ~~~,~~~~
  {\bf \dot d} = \frac{\negr d - \negr b}{||\negr d - \negr b||}
                 {\dot \rho}_2 =
  \left[
    \begin{array}{c}
      \cos(\alpha_2) \\
      \sin(\alpha_2)
    \end{array}
  \right] {\dot \rho}_2 \nonumber
  \eeqa
The two unactuated joint rates ${\dot \theta}_1$ and ${\dot
\theta}_2$ can be eliminated from equations (\ref{e:MGD_1}) and
(\ref{e:MGD_2}) by dot-multiplying the former  by $\negr p - \negr
c$ and the latter by $\negr p - \negr d$, thus obtaining
 \bseq
  \be
  (\negr p - \negr c)^T {\bf \dot p} = (\negr p - \negr c)^T
   \frac{\negr c - \negr a}{||\negr c - \negr a||}
  {\dot \rho}_1
  \ee
  \be
  (\negr p - \negr d)^T {\bf \dot p} = (\negr p - \negr d)^T
   \frac{\negr d - \negr b}{||\negr d - \negr b||}
  {\dot \rho}_2
  \ee
  \eseq
Equations (2a) and (2b) can now be cast in vector form, namely,
 \beqa
  \negr A {\bf \dot p} = \negr B \dot{\gbf \rho} \nonumber
 \eeqa
where \negr A and  \negr B are, respectively, the parallel and
serial Jacobian matrices, defined as
 \beqa
 \negr A = \left[
  \begin{array}{c}
    (\negr p - \negr c)^T \\
    (\negr p - \negr d)^T
  \end{array}
  \right] ~~~,~~~~ \negr B = \left[
  \begin{array}{cc}
    (\negr p - \negr c)^T
    ((\negr c - \negr a) /||\negr c - \negr a||)
    & 0 \\
    0 & (\negr p - \negr d)^T
    ((\negr d - \negr b) / ||\negr d - \negr b||)
  \end{array}
 \right]
 \nonumber
 \eeqa
and with $\dot{\gbf \rho}$ defined as the vector of actuated joint
rates and ${\bf \dot p}$ defined as the vector of velocity of
point $P$:
 \beqa
  {\dot{\gbf \rho}} =
  \left[
  \begin{array}{c}
    {\dot \rho}_1 \\
    {\dot \rho}_2
  \end{array}
 \right] ~~~&,&~~~~ {\bf \dot p} = \left[
  \begin{array}{c}
    \dot x \\
    \dot y
  \end{array}
 \right]
 \nonumber
 \eeqa
When \negr A and \negr B are not singular, we can study the
Jacobian kinematic  matrix \negr J \cite{Merlet:97},
 \bseq
 \be
 {\bf \dot p} = \negr J {\dot{\gbf \rho}} ~~~~{\rm with}~~~ \negr
 J= \negr A^{-1} \negr B
 \ee
or the inverse Jacobian kinematic matrix $\negr J^{-1}$, such that
 \be
  {\dot{\gbf \rho}} = \negr J^{-1} {\bf \dot p} ~~~~{\rm with}~~~
  \negr J^{-1}= \negr B^{-1} \negr A
  \ee
 \eseq
\subsection{Parallel Singularities}
The parallel singularities occur when the determinant of the
matrix \negr A vanishes \cite{Chablat:98,Gosselin:1990}, i.e. when
$\det(\negr A)=0$. In this configuration, it is possible to move
locally the tool center point whereas the actuated joints are
locked. These singularities are particularly undesirable, because
the structure cannot resist any force and control is lost. To
avoid any deterioration, it is necessary to eliminate the parallel
singularities from the workspace.

For the mechanism studied, the parallel singularities occur
whenever the points $C$, $D$, and $P$ are aligned
(Figure~\ref{figure:PRR_RP_Parallel_Singularity}), i.e. when
$\theta_1 - \theta_2 = k \pi$, for $k = 1,2,...$.
\begin{figure}[hbt]
    \begin{center}
    \begin{tabular}{cc}
       \begin{minipage}[t]{70 mm}
           \centerline{\hbox{\includegraphics
           [width= 69mm,height= 30mm]{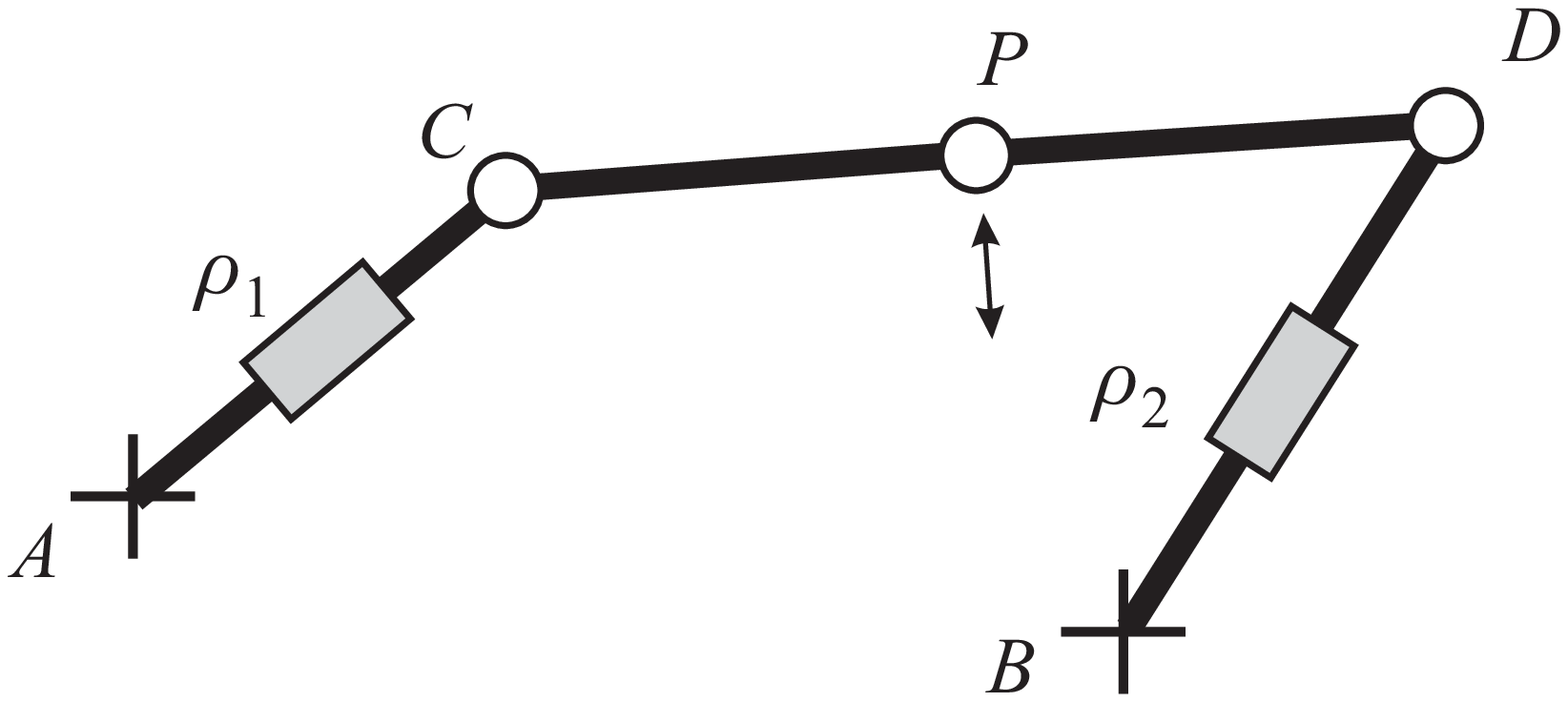}}}
           \caption{Parallel singularity}
           \protect\label{figure:PRR_RP_Parallel_Singularity}
       \end{minipage} &
       \begin{minipage}[t]{50 mm}
           \centerline{\hbox{\includegraphics
           [width= 43mm,height= 29mm]{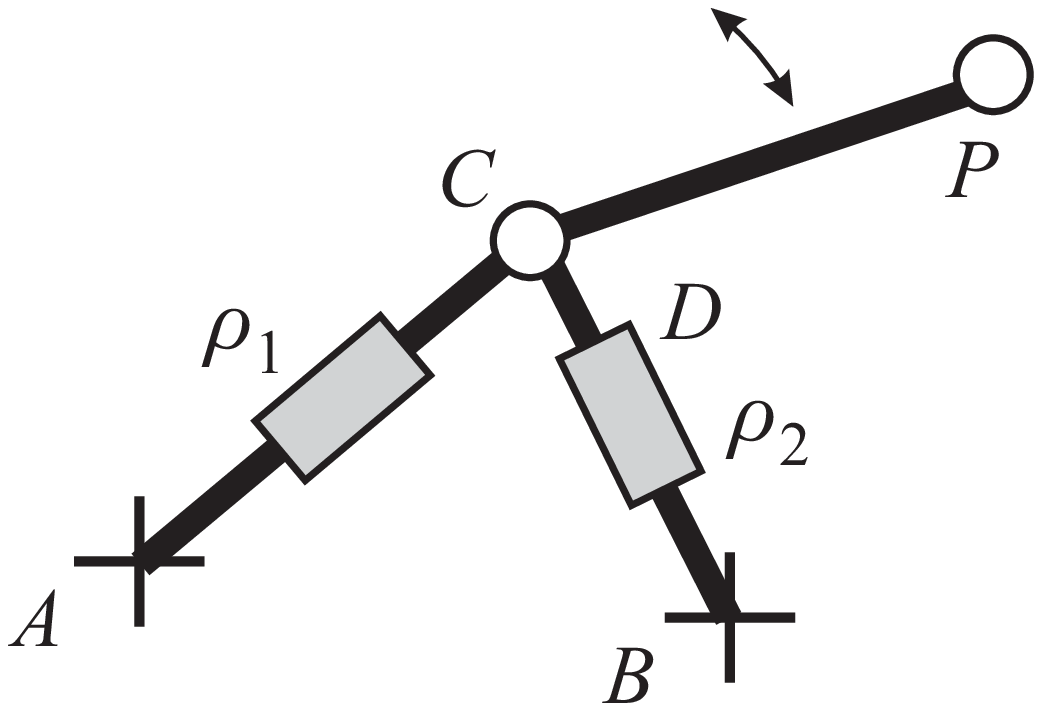}}}
           \caption{Structural singularity }
           \protect\label{figure:Structural_Singularity}
       \end{minipage}
    \end{tabular}
    \end{center}
\end{figure}
\par
They are located inside the Cartesian workspace and form the
boundaries of  the joint workspace. Moreover, structural
singularities can occur when $L_1$ is equal to $L_2$
(Figure~\ref{figure:Structural_Singularity}). In these
configurations,  the control of the point $P$ is lost.
\subsection{Serial Singularities}
The serial singularities occur when the determinant of the matrice
\negr B vanishes, i.e. when $\det(\negr B)=0$. When the
manipulator is in such configurations, there is a direction along
which no Cartesian velocity can be produced. The serial
singularities define the boundary of the Cartesian workspace
[Merlet~97].
 \begin{figure}[hbt]
    \begin{center}
        \includegraphics[width=65mm,height=31mm]{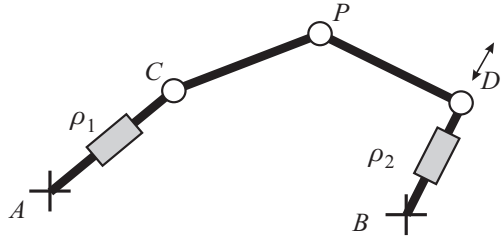}
        \caption{Serial singularity}
        \protect\label{figure:Serial_Singularity}
    \end{center}
 \end{figure}

For the topology studied, the serial singularities occur whenever
$\theta_1 - \alpha_1 =  \pi / 2 + k \pi$, or $\theta_2 - \alpha_2
= \pi / 2 + k\pi$, for $k = 1,2,...$
(Figure~\ref{figure:Serial_Singularity}), i.e whenever $AC$ is
orthogonal to $CP$ or $BD$ is orthogonal to $DP$.
\subsection{Application to Machining}
For a machine tool with three axes as in
(Figure~\ref{figure:machine_PPP}), the motion of the table is
 performed along two perpendicular axes. The joint limits of each
actuator give the dimension of the Cartesian workspace. For the
parallel mechanisms studied, this transformation is not direct.
The resulting Cartesian workspace is more complex and its size
smaller. We want to have a Cartesian workspace which will be close
to the Cartesian workspace of an industrial serial machine tool.
For our 2-DOF mechanisms, we will prescribe a rectangular shape
Cartesian workspace. In addition, the workspace must be reduced to
a t-connected region, i.e. a region free of serial and parallel
singularities \cite{Chablat:1999}. Finally, we want to prescribe
relatively stable kinetostatic properties in the workspace.

\subsection{Velocity Amplification Study}
In order to keep reasonable and homogeneous kinetostatic properties in
the Cartesian workspace, we study the manipulability ellipsoids of
velocity defined by the inverse Jacobian matrix $\negr J^{-1}$
\cite{Yoshikawa:85}. For the mechanisms at hand, the inverse
Kinematic Jacobian matrix $\negr J^{-1}$ given in equation
(3b) is simple. In this case, the matrices \negr B and $\negr
J^{-1}$ are written simply,
 \beqa
 \negr B= \frac{1}{L_1} \left[
  \begin{array}{cc}
      1 / c_1 & 0 \\
      0 & 1 / c_2
  \end{array}
 \right] ~~{\rm and}~~ \negr J^{-1}=
  \frac{1}{L_1}
  \left[
  \begin{array}{c}
      (1 / c_1) (\negr p - \negr c)^T \\
      (1 / c_2) (\negr p - \negr d)^T
  \end{array}
  \right] ~~{\rm with}~~ c_i= \cos(\theta_i - \alpha_i) ~~ i= 1, 2
 \nonumber
\eeqa

The square roots $\gamma_1$ and $\gamma_2$ of the eigenvalues of
 $(\negr J \negr J^T)^{-1}$ are the values of the semi-axes
 of the ellipse which define the two factors of velocity amplification
 (from the joint rates to the output velocities), $\lambda_1 = 1 / \gamma_1$ and
$\lambda_2 = 1 / \gamma_2$, according to these principal  axes. To
limit the variations of this factor in the Cartesian workspace, we
pose the following constraints,
 \beqa
  1/3 < \lambda_i < 3
  \label{equation:contraintes}
 \eeqa
This means that for a given joint velocity, the output velocity is
either at most three times larger or, at least, three times
smaller. This constraint also permits to limit the loss of
rigidity (velocity amplification lowers rigidity) and of accuracy
(velocity amplification also amplifies the encoder resolution).
The values in equation (4) were chosen as an example and should be
defined precisely as a function of the type of machining tasks.
\section{Comparative Study}
\subsection{The Three Parallel Mechanism Architectures Studied}
The three parallel mechanism architectures studied are the following:
\begin{enumerate}
 \item The biglide1 mechanism with $\alpha_1= 0$ and $\alpha_2= \pi$
 (Fig.~\ref{figure:1_mechanism})
 \item The biglide2 mechanism with $\alpha_1= \pi / 2$ and $\alpha_2= \pi / 2$
 (Fig.~\ref{figure:2_mechanism})
 \item The orthoglide mechanism with $\alpha_1= \pi / 4$ and $\alpha_2= 3\pi /4$
 (Fig.~\ref{figure:3_mechanism})
\end{enumerate}

The biglide1 mechanism studied (Fig.~\ref{figure:1_mechanism}) has
been used for example in the hexaglide and in the triglide
\cite{Wenger:2000}.

\begin{figure}[!ht]
    \begin{center}
           \includegraphics[width=56mm,height=31mm]{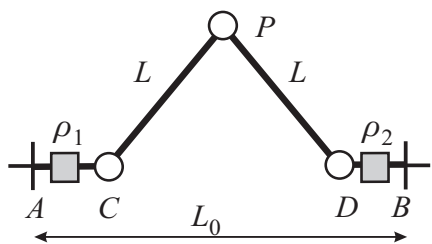}
           \caption{The biglide1 mechanism}
           \protect\label{figure:1_mechanism}
    \end{center}
\end{figure}

The biglide2 mechanism (Fig.~\ref{figure:2_mechanism}) has been
used in the Linapod and in \cite{Wenger:2000, Horn:2000}.

\begin{figure}[!ht]
    \begin{center}
    \begin{tabular}{cc}
       \begin{minipage}[t]{60 mm}
           \centerline{\hbox{\includegraphics[width=30mm,height=46mm]{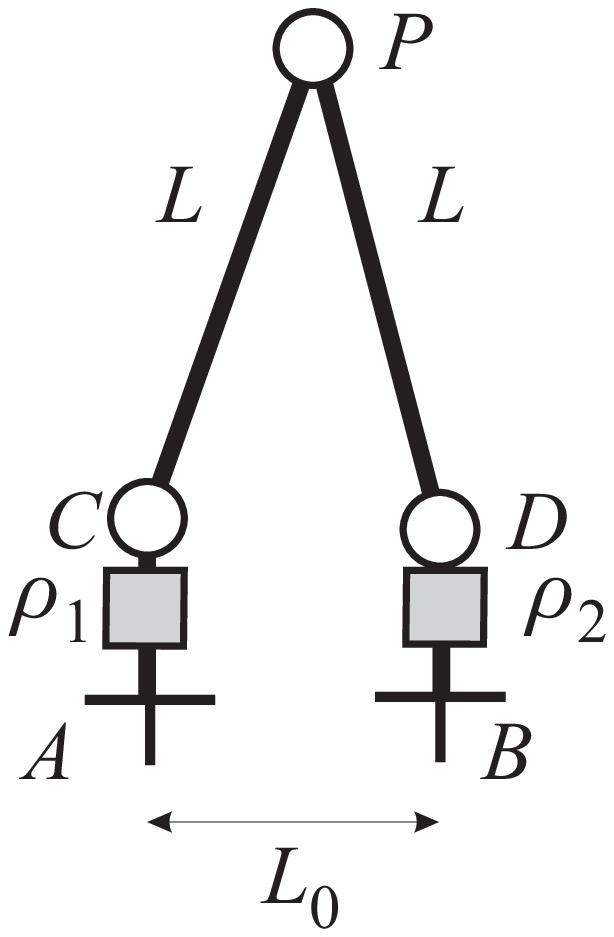}}}
           \caption{The biglide2 mechanism}
           \protect\label{figure:2_mechanism}
       \end{minipage} &
       \begin{minipage}[t]{60 mm}
           \centerline{\hbox{\includegraphics[width=57mm,height=39mm]{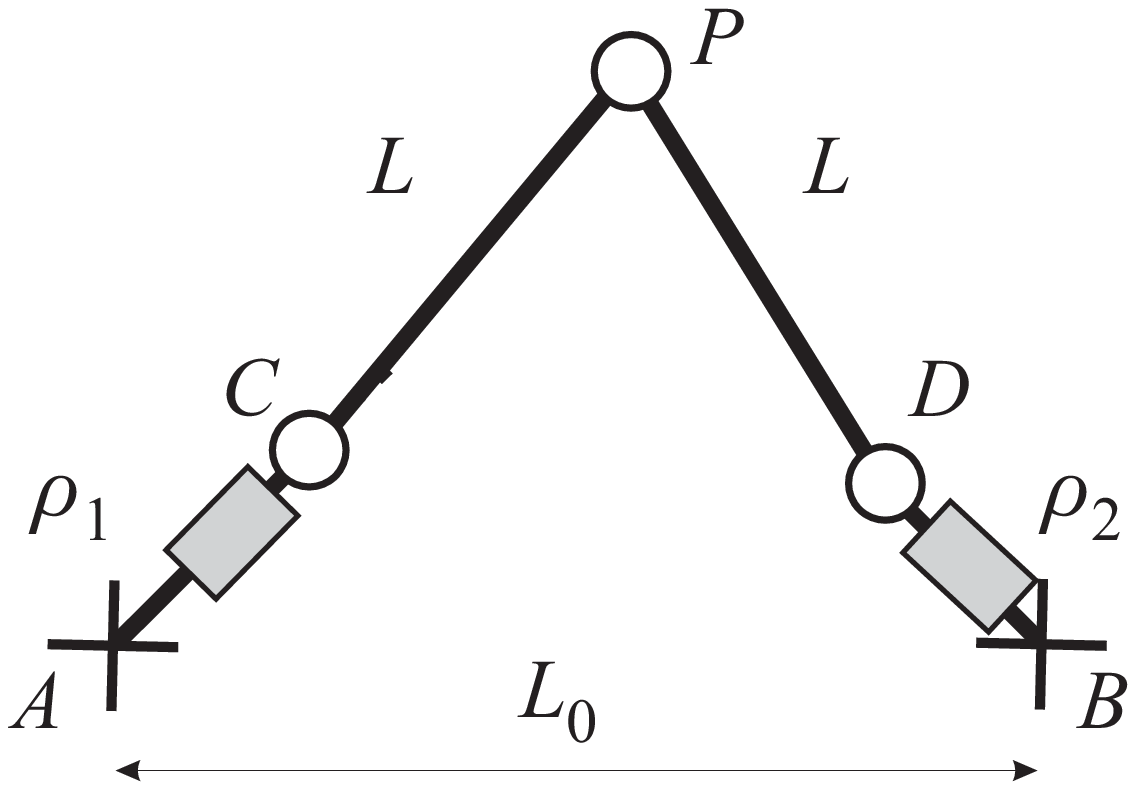}}}
           \caption{The orthoglide mechanism}
           \protect\label{figure:3_mechanism}
       \end{minipage}
    \end{tabular}
    \end{center}
\end{figure}

The third mechanism (Fig.~\ref{figure:3_mechanism}) was introduced
in \cite{Chablat:2000} and extended to 3-DOF in
\cite{Wenger:2000}. The main constraint of this design is $AC
\perp BD$, which makes it isotropic, i.e. the Jacobian matrix of
this mechanism is isotropic in some configurations.
\subsection{Determination of the Mechanism Dimensions}
To determine the mechanism dimensions, we proceed in several steps
as follows. Let $L = L_1 = L_2$ be the common link lengths, let
$L_0$ be the distance between the attachment points A and B of the
prismatic joints and let $\Delta \rho$ be the range of the
actuated joints. The length $L_0$ and the joint range $\Delta
\rho$ are determined consecutively as function of $L$ by using the
condition on the velocity amplification factor (see section 2.7).
For all mechanisms, we can show that the maximal (resp. minimal)
velocity amplification factor is reached at the configuration for
which the distance between C and D is maximal (resp. minimal). For
the biglide1 and for the orthoglide, the maximal (resp. minimal)
velocity amplification factor is reached at the configuration
where C is on A and D is on B (resp. where C is on C' and D is on
D') (figures~\ref{figure:10_mechanism} and
\ref{figure:30_mechanism}).

\begin{figure}[!ht]
    \begin{center}
           \includegraphics[width=57mm,height=37mm]{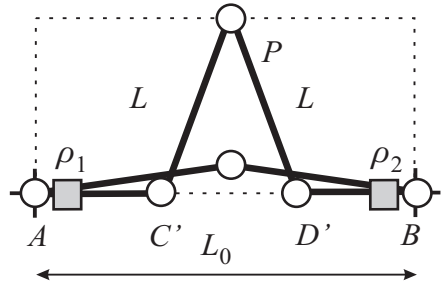}
           \caption{The biglide1 mechanism}
           \protect\label{figure:10_mechanism}
    \end{center}
\end{figure}
\begin{figure}[!ht]
    \begin{center}
    \begin{tabular}{cc}
       \begin{minipage}[t]{60 mm}
           \centerline{\hbox{\includegraphics[width=36mm,height=53mm]{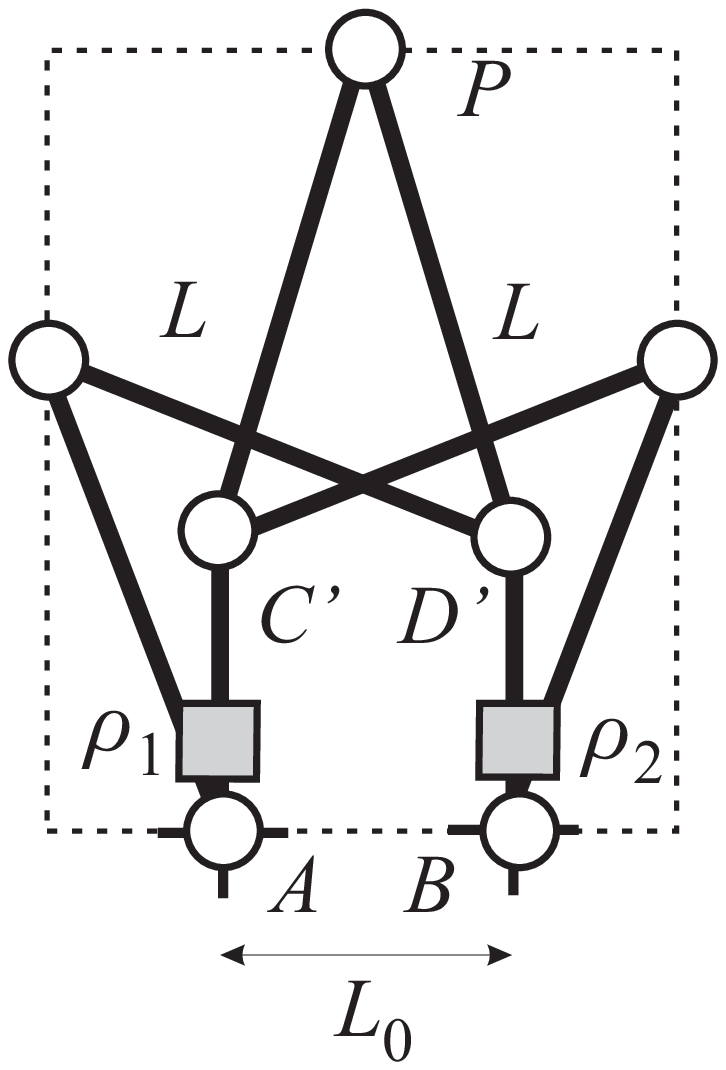}}}
           \caption{The biglide2 mechanism}
           \protect\label{figure:20_mechanism}
       \end{minipage} &
       \begin{minipage}[t]{60 mm}
           \centerline{\hbox{\includegraphics[width=58mm,height=45mm]{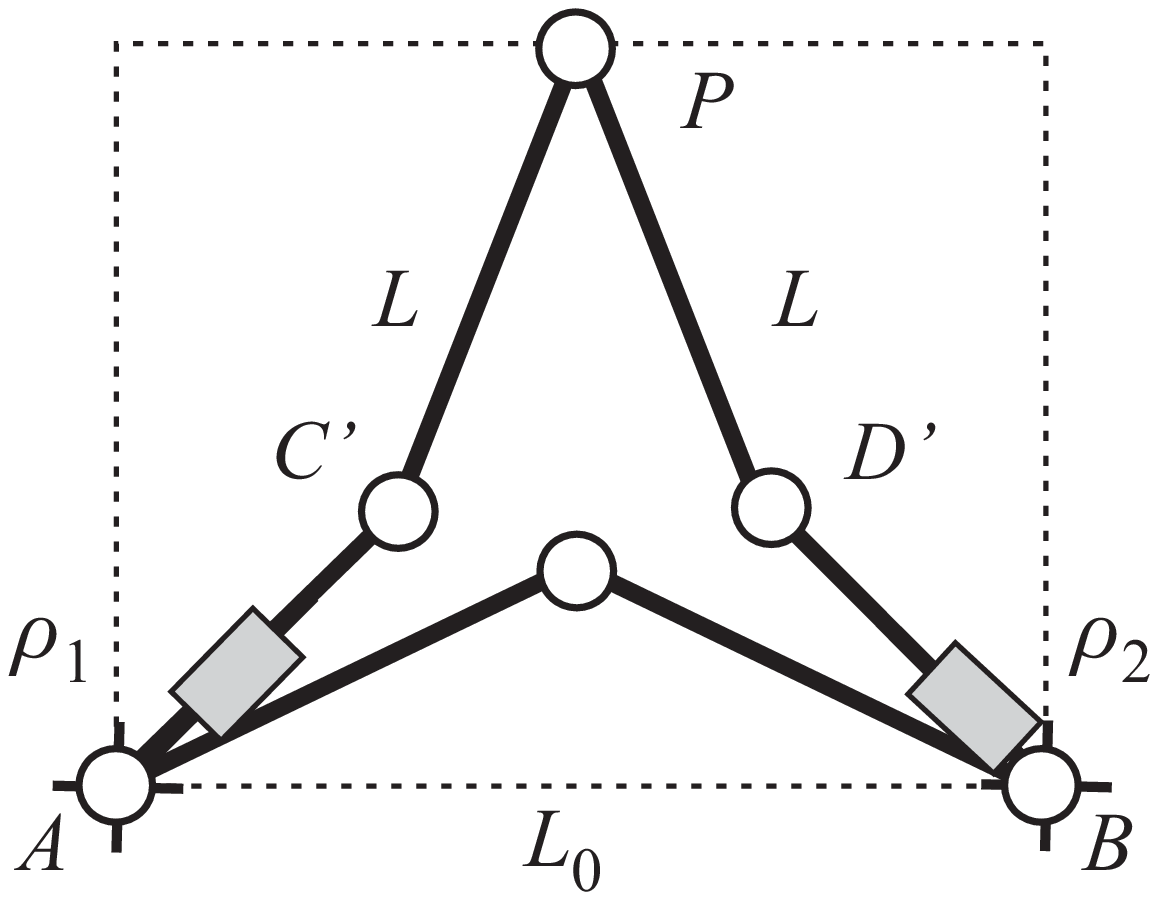}}}
           \caption{The orthoglide mechanism}
           \protect\label{figure:30_mechanism}
       \end{minipage}
    \end{tabular}
    \end{center}
\end{figure}

By first writing that the maximal factor must be smaller than 3 in
the first configuration, we can calculate $L_0$. Then $\Delta
\rho$ is calculated by writing that in the opposite configuration
the velocity amplification factor must be larger than 1/3. For the
biglide2, the maximal (resp. minimal) amplification factor is
reached at the configuration where C is on A and D is on D'(resp.
where C and D lie on an horizontal line)
(figure~\ref{figure:20_mechanism}). In this case, we first
calculate $L_0$ at the minimal factor configuration and $\Delta
\rho$ is then calculated at the maximal factor configuration. The
values of $L_0$ and $\Delta \rho$ obtained for all mechanisms are
given in the first two rows of table~\ref{Table:L_1}. All
derivations and computations have been obtained with MAPLE.

\begin{table}[!ht]
  \begin{center}
     \begin{tabular}{|c|c|c|c|c|} \hline
       Mechanism  & $L_0$    & $\Delta \rho$  & $S$      \\ \hline
       Biglide1   & $1.946L$ & $0.547L$       & $0.107L$ \\ \hline 
       Biglide2   & $0.458L$ & $0.529L$       & $0.249L$ \\ \hline 
       Orthoglide & $1.961L$ & $1.109L$       & $0.885L$ \\ \hline 
     \end{tabular}
  \caption{Dimensions and rectangular Cartesian workspace surface as function of $L$}
  \label{Table:L_1}
  \end{center}
\end{table}

Then, for each mechanism, we determine the maximum rectangular
surface $S$ which can be included in the Cartesian workspace
(figures \ref{figure:WCK_1_A} to \ref{figure:WCK_4_A}). We have
used the parametric sketcher of a CAD system to perform this task.
The area of the surfaces $S$ obtained are given in the last row of
table~\ref{Table:L_1}. The last step is the scaling of the
mechanism link dimensions and joint ranges in order to have a same
Cartesian workspace rectangular for all mechanisms. The first
three rows of table~\ref{Table:Optimized_Lengths} give the
resulting link dimensions and joint ranges.

\begin{figure}[!ht]
    \begin{center}
    \begin{tabular}{ccc}
       \begin{minipage}[t]{45 mm}
        \centerline{\hbox{\includegraphics[width=31mm,height=18mm]{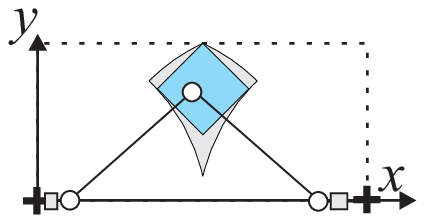}}}
        \caption{The Cartesian workspace of the Biglide1}
        \protect\label{figure:WCK_1_A}
       \end{minipage} &
       \begin{minipage}[t]{45 mm}
        \centerline{\hbox{\includegraphics[width=25mm,height=24mm]{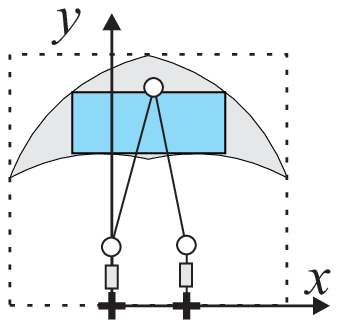}}}
        \caption{The Cartesian workspace of the Biglide2}
        \protect\label{figure:WCK_2_A}
       \end{minipage} &
       \begin{minipage}[t]{45 mm}
        \centerline{\hbox{\includegraphics[width=36mm,height=30mm]{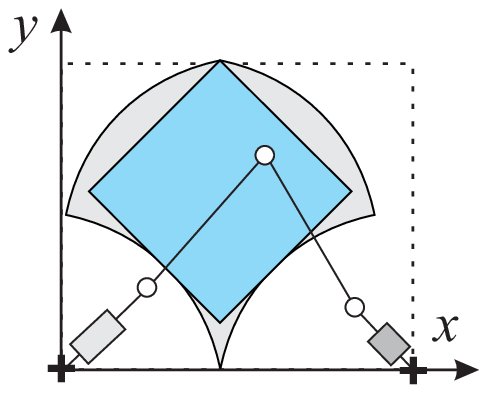}}}
        \caption{The Cartesian workspace of the Orthoglide}
        \protect\label{figure:WCK_4_A}
       \end{minipage}
    \end{tabular}
    \end{center}
\end{figure}

\subsection{Comparison of the Mechanism Size Envelopes}
Table~\ref{Table:Optimized_Lengths} provides the mechanism
dimensions and envelope sizes for the three parallel mechanisms
studied, for a prescribed rectangular Cartesian workspace surface
of $1~m^2$.
\begin{table}[!ht]
  \begin{center}
     \begin{tabular}{|c||c|c|c|c|c|c|c|c|c|c|c|c|} \hline
       Mechanism   & $L_0$   & $L$    &$\Delta \rho$  & Mechanism envelope surface      \\ \hline
       Biglide1    & $5.95$ & $3.05$ & $1.67$       & $16.45$ \\ \hline
       Biglide2    & $0.92$ & $2.00$ & $1.06$       & $ 8.50$ \\ \hline
       Orthoglide  & $2.08$ & $1.06$ & $1.18$       & $ 3.91$ \\ \hline
     \end{tabular}
  \caption{Mechanism dimensions and envelope sizes for a same rectangular Cartesian
workspace}
  \label{Table:Optimized_Lengths}
  \end{center}
\end{table}

Figs.~\ref{figure:Workspace_1}, \ref{figure:Workspace_2} and
\ref{figure:Workspace_3} show the three mechanisms along with
their Cartesian workspace and the same rectangular surface in it.
We can notice that the orthoglide mechanism has smaller lengths
struts  i.e. smaller mass in motion and thus higher dynamic
performances than the other two mechanisms. The biglide2 and the
orthoglide mechanisms have similar values of $\Delta \rho$. It
should be noticed, also, that the Cartesian workspace of the
biglide2 includes a rectangulle which is far from a square,
whereas it is an exact square for the other two mechanisms. We
have calculated the dimensions of the biglide2 for a square of
$1~m^2$ in its workspace and we have obtained $L_0=1.075$,
$L=2.348$ and $\Delta \rho=1.242$.

\begin{figure}[!ht]
    \begin{center}
        \includegraphics[width=87mm,height=40mm]{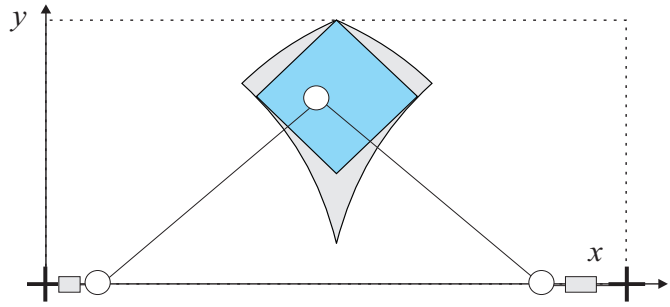}
        \caption{Workspace of the biglide1 mechanism}
        \protect\label{figure:Workspace_1}
    \end{center}
\end{figure}

\begin{figure}[!ht]
    \begin{center}
    \begin{tabular}{cc}
       \begin{minipage}[t]{65 mm}
           \centerline{\hbox{\includegraphics[width=48mm,height=46mm]{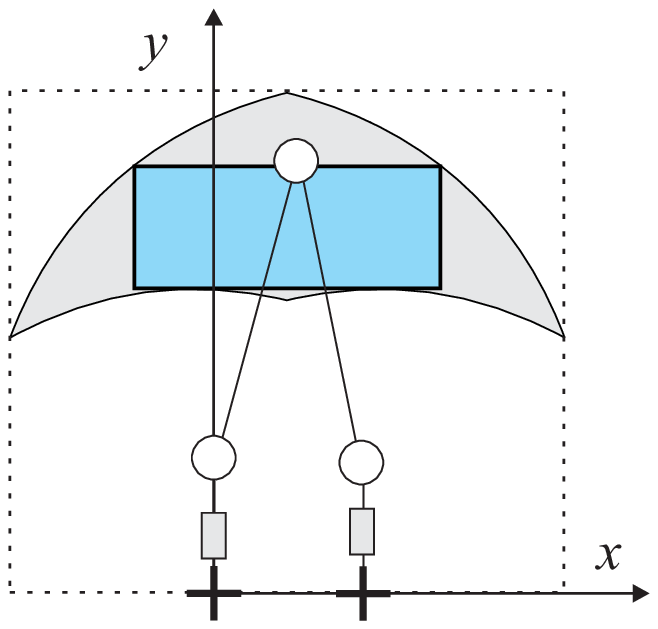}}}
           \caption{Workspace of the biglide2 mechanism}
           \protect\label{figure:Workspace_2}
       \end{minipage} &
       \begin{minipage}[t]{65 mm}
           \centerline{\hbox{\includegraphics[width=40mm,height=31mm]{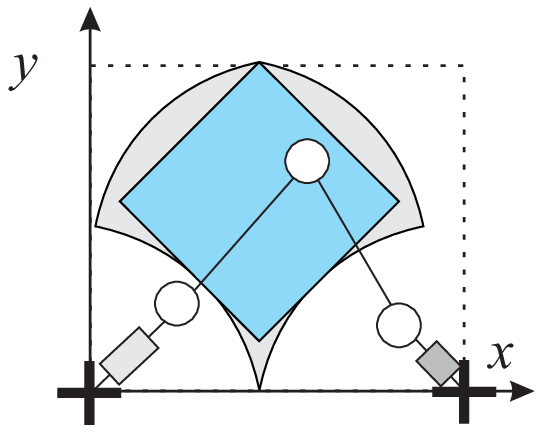}}}
           \caption{Workspace of the orthoglide mechanism}
           \protect\label{figure:Workspace_3}
       \end{minipage}
    \end{tabular}
    \end{center}
\end{figure}

\section{Conclusions}
Three 2-DOF parallel mechanisms dedicated to machining
applications have been compared in this paper. The link dimensions
and the actuated joint ranges have been calculated for a same
prescribed rectangular Cartesian workspace with identical
kinetostatic constraints. The machine size of each resulting
design was used as a comparative criterion. One of the mechanisms,
the orthoglide, was shown to have lower dimensions than the other
two mechanisms. This result shows that the isotropic property of
the orthoglide induces interesting additional features like better
compactness and lower inertia. In the future, the comparative
study will be continued using dynamic performance indices.
\bibliographystyle{unsrt}

\end{document}